\documentclass[10pt,twocolumn,letterpaper]{article}

\usepackage{iccv}
\usepackage{times}
\usepackage{epsfig}
\usepackage{graphicx}
\usepackage{amsmath}
\usepackage{amssymb}
\usepackage{algorithm}
\usepackage{algorithmic}
\usepackage{bbm}
\usepackage{amsfonts}
\usepackage{mathrsfs}
\usepackage{multirow}
\usepackage{bm}

\usepackage[pagebackref=true,breaklinks=true,letterpaper=true,colorlinks,bookmarks=false]{hyperref}

\iccvfinalcopy 

\begin{document}

\title{Dual Cross-Attention for Medical Image Segmentation}

\author{Gorkem Can Ates{$^{1*}$},~ Prasoon Mohan{$^{1}$},~ Emrah Celik{$^{1}$}\\\vspace{-8pt}{\small~}\\
{$^{1}$}University of Miami\\
{\small{{$^{*}$}Corresponding to: \tt{gca45@miami.edu}}}
}

\maketitle
\thispagestyle{empty}

\begin{abstract}

We propose Dual Cross-Attention (DCA), a simple yet effective attention module that is able to enhance skip-connections in U-Net-based architectures for medical image segmentation. DCA addresses the semantic gap between encoder and decoder features by sequentially capturing channel and spatial dependencies across multi-scale encoder features. First, the Channel Cross-Attention (CCA) extracts global channel-wise dependencies by utilizing cross-attention across channel tokens of multi-scale encoder features. Then, the Spatial Cross-Attention (SCA) module performs cross-attention to capture spatial dependencies across spatial tokens. Finally, these fine-grained encoder features are up-sampled and connected to their corresponding decoder parts to form the skip-connection scheme. Our proposed DCA module can be integrated into any encoder-decoder architecture with skip-connections such as  U-Net and its variants. We test our DCA module by integrating it into six U-Net-based architectures such as U-Net, V-Net, R2Unet, ResUnet++, DoubleUnet and MultiResUnet. Our DCA module shows Dice Score improvements up to 2.05\% on GlaS, 2.74\% on MoNuSeg, 1.37\% on CVC-ClinicDB, 1.12\%  on Kvasir-Seg and 1.44\% on Synapse datasets. Our codes are available at: \url{https://github.com/gorkemcanates/Dual-Cross-Attention}

\end{abstract}
\section{Introduction}
\label{sec:introduction}
\begin{figure*}[t]
\centering
\includegraphics[width=15.5cm]{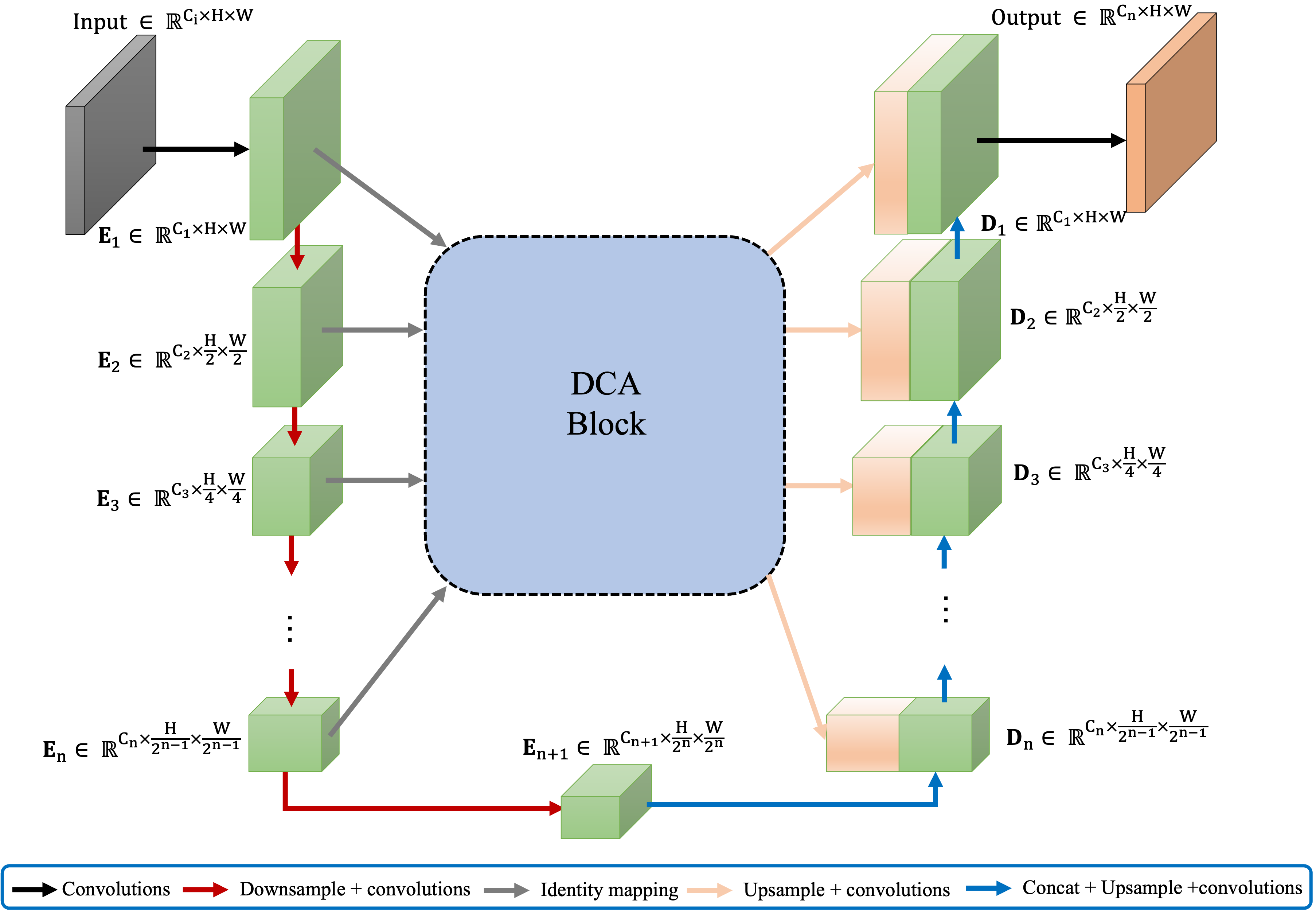}\\
\caption{Encoder decoder architecture with our proposed DCA block. DCA block can be integrated into any encoder-decoder architecture with skip connections. It takes multi-scale features from different encoder stages, produces enhanced representations and connects them to their decoder counterparts.}
\label{fig:1} 
\end{figure*}

Convolutional Neural Networks (CNNs) have become the de-facto standard for accurate medical image segmentation because of their strong and complex mapping capability \cite{gu2018recent}. Fully Convolutional Networks (FCNs) \cite{long2015fully}, particularly convolutional encoder-decoder networks have drawn much attention in the past years due to their tremendous success in various medical image segmentation tasks \cite{wang2015unified, hu2017automatic}. U-Net \cite{ronneberger2015u}, in particular, achieved superior performance due to its skip-connection scheme, which connects low-level features extracted by the encoder to the decoder to achieve a better feature representation. Such a scheme helps the model recover the contextual information loss during the down-sampling process in the encoder by simply concatenating encoder features on different scales with their corresponding parts in the decoder. Motivated by the success of U-Net and the skip-connection scheme, several architectural designs have been developed \cite{cciccek20163d, milletari2016v, zhou2018unet++, alom2018recurrent, schlemper2019attention, jha2019resunet++, jha2020doubleu}. These U-Net variants have successfully improved vanilla U-Net in various medical image segmentation tasks by adapting sophisticated designs such as residual \cite{he2016deep} and recurrent \cite{liang2015recurrent} connections into the encoder-decoder framework and/or improving the plain skip-connection scheme through further enhancing the encoder features before connecting to the decoder.

Although U-Net as well as its variants achieved good performance on various medical image segmentation tasks, there still exist performance limitations. The first limitation comes from the locality of convolutions which cannot capture the long-range dependencies across different features. This is mainly caused by the nature of the convolutional operation which gradually obtains the local receptive fields using local kernels rather than extracting global feature interactions at once \cite{hu2019local}. The second limitation is the semantic gap caused by skip-connections when simply concatenating encoder and decoder features. Recently, Wang \textit{et al.} \cite{wang2022uctransnet} showed that the plain skip-connection scheme presented by U-Net is not sufficient enough to model the global multi-scale context and it is in fact essential to effectively fuse the low-level encoder features before connecting them to their corresponding decoder parts. As mentioned before, several U-Net variations tackled such a semantic gap problem by adding a series of convolutional or residual layers to further fuse the low-level encoder and decoder features in order to adequately connect them to their decoder counterparts. One approach is U-Net++ \cite{zhou2018unet++}, which connects encoder features to decoder features through a series of nested, dense skip pathways. Another approach is MultiResUnet \cite{ibtehaz2020multiresunet}, which introduced residual paths by applying a series of convolutional layers with residual connections on encoder features propagating to their decoder counterparts. Despite improving the quality of the skip-connections, both of these methods still struggle to reduce the semantic gap between encoder and decoder features.

Recently, Transformers \cite{vaswani2017attention}, originally proposed for Natural Language Processing (NLP), have become the dominant architecture in computer vision with the pioneering work of Vision Transformer (ViT) \cite{dosovitskiy2020image}. Subsequently, its success has been shown in various vision tasks including image classification \cite{liu2021swin, touvron2021training, yang2021focal, wang2021pyramid, huang2021shuffle, dong2022cswin, lee2022mpvit}, object detection \cite{carion2020end, fang2021you, wang2022anchor}, segmentation \cite{strudel2021segmenter, xie2021segformer, zheng2021rethinking, guo2021sotr} and beyond \cite{wang2021sceneformer, chen2021pre, yang2020learning, li2019visualbert, sun2019videobert, neimark2021video, wang2021end}. The self-attention mechanism in transformers certainly plays a key role in their success due to its ability to directly capture long-range dependencies \cite{ramachandran2019stand}. Motivated by the effectiveness of self-attention, researchers proposed to combine channel self-attention with self-attention to also capture the channel-wise interactions. These dual attention schemes based on self-attention have been shown to improve performance on various vision tasks. For instance, Fu \textit{et al.} \cite{fu2019dual} proposed DANet for scene segmentation, where a channel and a spatial attention module were integrated at the end of a dilated FCN to model the semantic dependencies in both spatial and channel dimensions. Those two attention modules were applied in a parallel manner to capture channel and spatial dependencies separately. Then, those two outputs were fused using element-wise summation to further improve the feature representations, which significantly improved the segmentation performance. Mou \textit{et al.} \cite{mou2021cs2} performed a similar fusing strategy by summing the channel and spatial attention in a convolutional encoder-decoder network for the segmentation of curvilinear structures. Liu \textit{et al.} \cite{liu2021scsa}, on the other hand, used concatenation to fuse the outputs of channel and spatial attention modules. A recent study \cite{ding2022davit} proposed a different dual attention approach named DaViT, in which spatial and channel attention modules were utilized sequentially. That is, self-attention is first applied to the spatial tokens, then the outputs of the spatial attention module are fed into the channel attention module which generates the fine-grained features by further extracting long-range feature interactions. Such dual attention mechanisms showed that the channel-self attention mechanism can also provide useful information. In fact, Wang \textit{et al.} \cite{wang2022uctransnet} showed that even stand-alone channel-wise attention can effectively capture the global context. They introduced channel cross-attention to tackle the semantic gap problem in U-Net by utilizing cross-attention in the channel axis of the multi-scale encoder features to capture the long-range channel dependencies. Their cross-attention mechanism improved the segmentation performance on different medical image segmentation datasets when utilized with the U-Net structure, showing its promise in reducing the semantic gap between encoder and decoder features. 

Motivated by the success of sequential dual attention \cite{ding2022davit} and the channel cross-attention \cite{wang2022uctransnet}, we propose Dual Cross-Attention (DCA), an attention module that effectively extracts channel and spatial-wise interdependencies across multi-scale encoder features to tackle the semantic gap problem. A major difference between channel and spatial cross-attention mechanisms is that channel-cross attention extracts global channel-wise context by fusing all spatial positions together between any given two channels, whereas spatial-cross attention mechanism extracts global spatial context by capturing spatial interdependencies between any two positions of the multi-scale encoder channels. A delicate incorporation of such channel and spatial cross-attention modules can further extract long-range contextual information by capturing the rich global context in both channel and spatial dimensions. Note that our main purpose here is to develop a well-structured mechanism that acts as a powerful bridge between encoder and decoder at a slight parameter increase rather than building an end-to-end network with an extensive amount of parameters as in \cite{chen2021transunet}. Note that, MetaFormer \cite{yu2022metaformer} the general, abstracted architecture in end-to-end transformers consists of a patch embedding operation (e.g., convolutions) \cite{xiao2021early, yuan2021tokens, wu2021cvt, wang2022convolutional} and two main building blocks, namely a token mixer (e.g., attention \cite{vaswani2017attention}) and a two-layer multi-layer perceptron (MLP) block with a non-linear activation (e.g., GeLU \cite{hendrycks2016gaussian}) \cite{yu2022metaformer2}. All of these blocks typically come with large computational costs, especially when multiple MetaFormer blocks are stacked together to form an end-to-end vision transformer network \cite{dosovitskiy2020image, liu2021swin, tolstikhin2021mlp, touvron2022resmlp, touvron2021training, yu2022metaformer, yang2022focal}. We minimize such computational costs by first introducing a new patch embedding operation, where we extract tokens by utilizing 2D average pooling which does not require any additional parameters and project them with $1\times1$ depth-wise convolutions. Then, we replace the linear projections in both channel and spatial attention modules with depth-wise convolutions \cite{chollet2017xception}. Finally, we completely eliminate the MLP layer to further reduce the number of parameters and we up-sample the DCA outputs to connect them to the decoder, thus forming our DCA mechanism. As we will show later, such modifications can still improve the segmentation performance while minimizing the computational overhead that occurs in MetaFormer. Extensive experiments using six state-of-the-art U-Net-based architectures and five benchmark medical image segmentation datasets show that our DCA module can significantly improve segmentation performance with minimal computational overhead.

\section{Dual Cross-Attention (DCA)}
\label{sec:dualcrossattention}
\begin{figure*}[t]
\centering
\includegraphics[width=14.3cm, height=9.7cm]{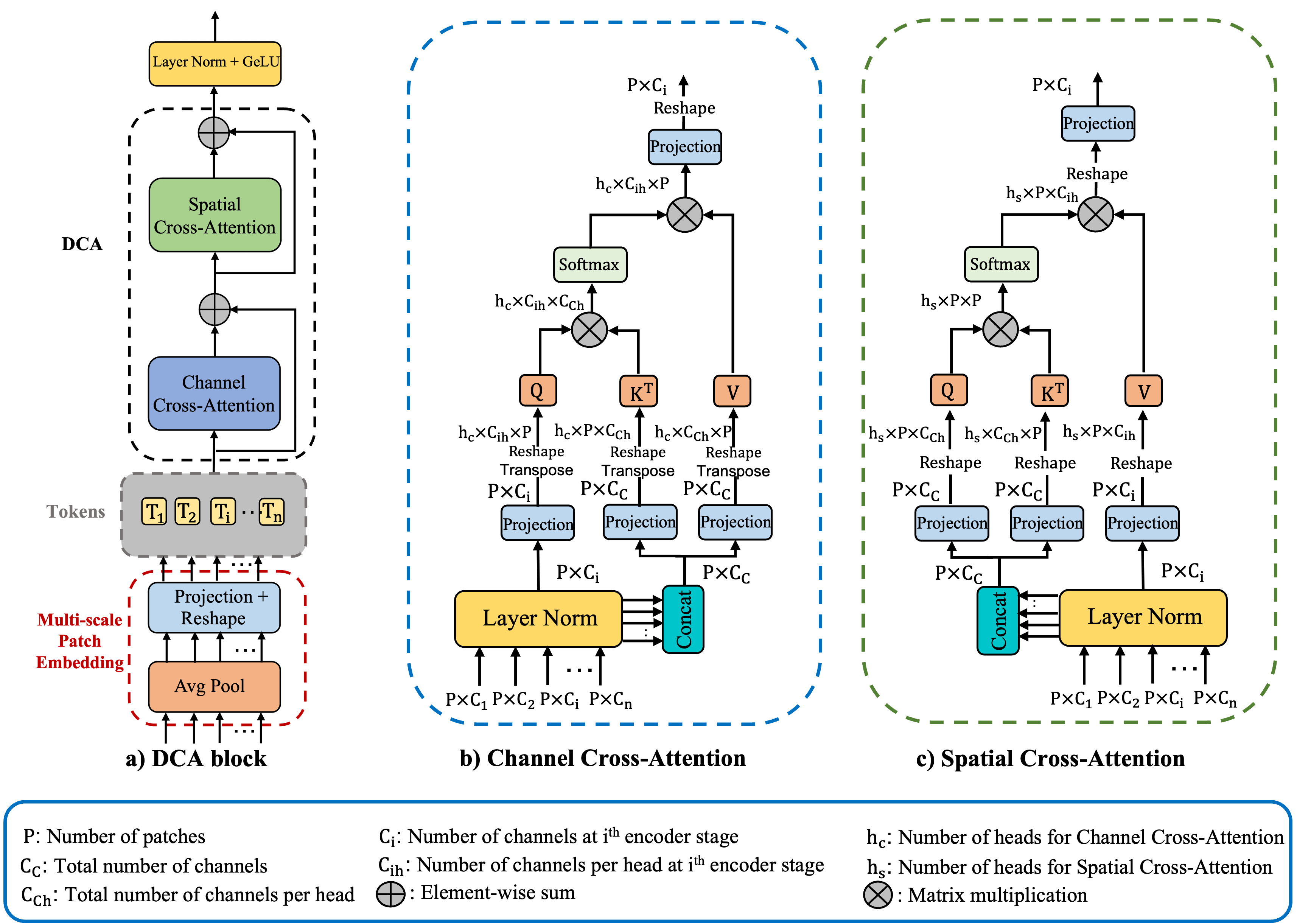}\\
\caption{Architecture of our proposed DCA block (\textbf{a)}). It consists of \textbf{b)} Channel Cross-Attention and \textbf{c)} Spatial Cross-Attention modules to capture long-range interactions.}
\label{fig:2} 
\end{figure*}

Fig.\ref{fig:1} illustrates the integration of our proposed DCA block into a general FCN architecture with skip-connections. The architecture of our DCA block is invariant to the number of encoder stages. That is, given n+1 multi-scale encoder stages, the DCA block takes multi-scale features  from the first n stages as input (outputs of the last convolutional layers in each stage), produces enhanced representations, and connects them to their corresponding n decoder stages. As shown in Fig.\ref{fig:2}a, we divide our DCA block into two main stages. The first stage consists of a multi-scale patch embedding module to obtain encoder tokens. In the second stage, we perform our proposed DCA mechanism using channel cross-attention (CCA) and spatial cross-attention (SCA) modules on these encoder tokens to capture long-range dependencies. Finally, we apply layer normalization \cite{ba2016layer} and GeLU \cite{hendrycks2016gaussian} sequence and upsample those tokens to connect them to their decoder counterparts.

\subsection{Patch Embedding from Multi-Scale Encoder Stages}
We start by extracting patches from n multi-scale encoder stages (i.e., skip-connection layers). Given n encoder stages from different scales $\pmb{\text{E}}_\text{i} \in \mathbb{R} ^ {\text{C}_\text{i}\times\frac{\text{H}}{\text{$2^\text{i-1}$}}\times\frac{\text{W}}{\text{$2^\text{i-1}$}}}$ and patch sizes $\text{P}^s_\text{i} = \frac{P^s}{2^{i-1}}$ where i = 1, 2, ..., n, we extract patches using 2D average pooling where the pool size and stride are $\text{P}^s_\text{i}$ and apply projection using $1\times1$ depth-wise convolutions over the flattened 2D patches: 
\begin{equation}
\textbf{T}_\text{i} = \text{DConv1D}_{\textbf{E}_\text{i}}(\text{Reshape}(\text{AvgPool2D}_{\textbf{E}_\text{i}}(\textbf{E}_\text{i})))
\end{equation}

where $\textbf{T}_\text{i} \in \mathbb{R} ^ {\text{P}\times\text{C}_\text{i}}$, (i = 1, 2, ..., n) represents the flattened patches for the $\text{i}^{\text{th}}$ encoder stage. Note that P is the number of patches which is the same for each $\pmb{\text{T}}_\text{i}$ so that we can utilize cross-attention across those tokens.

\subsection{Channel Cross-Attention (CCA)}
As shown in Fig.\ref{fig:2}b, each token $\pmb{\text{T}}_\text{i}$ is fed into the CCA module. We first perform layer normalization (LN) \cite{ba2016layer} on each $\pmb{\text{T}}_\text{i}$. Then, by following the cross-attention strategy in \cite{wang2022uctransnet}, we concatenate tokens $\pmb{\text{T}}_\text{i}$, (i = 1, 2, ..., n) along the channel dimension the create our keys and values $\pmb{\text{T}}_\text{c}$ while we use $\pmb{\text{T}}_\text{i}$ for the queries. Although linear projection is typically utilized in conventional self-attention, recent studies successfully adapted convolutions into self-attention in order to bring locality as well as to reduce computational complexity \cite{wang2018non, wu2021cvt, wang2021evolving}. Depth-wise convolutions, in particular, have been utilized in self-attention because of their ability to capture local information with negligible additional computational cost \cite{xu2021co, lu2021container, guo2022cmt, chen2022mixformer, lee2022mpvit}. Motivated by prior work and such advantages, we replace all linear projections with $1\times1$ depth-wise convolutional projections: 

\begin{equation}
    \textbf{Q}_\text{i} = \text{DConv1D}_{\textbf{Q}_\text{i}}(\textbf{T}_\text{i}) 
\end{equation}
\begin{equation}
        \textbf{K} = \text{DConv1D}_{\textbf{K}}(\textbf{T}_\text{c}) 
\end{equation}
\begin{equation}
        \textbf{V} = \text{DConv1D}_{\textbf{V}}(\textbf{T}_\text{c})
\end{equation}

where $\pmb{\text{Q}}_\text{i} \in \mathbb{R} ^ {\text{P}\times\text{C}_\text{i}}$, $\pmb{\text{K}} \in \mathbb{R} ^ {\text{P}\times\text{C}_\text{c}}$, $\pmb{\text{V}} \in \mathbb{R} ^ {\text{P}\times\text{C}_\text{c}}$ are the projected queries, keys, and values, respectively. In order to utilize cross-attention along the channel dimension, we take the transpose of queries, keys, and values. Thus, CCA takes the following form:
\begin{equation}
\text{CCA}(\textbf{Q}_\text{i}, \textbf{K}, \textbf{V}) = \text{Softmax}\left(\frac{{\textbf{Q}_\text{i}}^T\textbf{K}}{\sqrt{\textbf{C}_\text{c}}}\right)\textbf{V}^T
\end{equation}

where $\textbf{Q}_\text{i}$, $\textbf{K}$, and $\textbf{V}$ are matrices representing the queries, keys, and values, respectively, and $\frac{1}{\sqrt{\textbf{C}_\text{c}}}$ is the scaling factor. The output of cross-attention is a weighted sum of the values, where the weights are determined by the similarity between the queries and keys. The output of the softmax function is then used to weight the values. Finally, we apply depth-wise convolutional projections to the outputs of the cross-attention and feed them into the SCA module.

\begin{table*}[t]
\scriptsize
\renewcommand\arraystretch{1.4}
\begin{center}
    
\begin{tabular}{|cccccccccccc|}
\hline
\multicolumn{2}{|l}{}                                                           & \multicolumn{2}{c}{\textbf{GlaS}}                 & \multicolumn{2}{c}{\textbf{MoNuSeg}}              & \multicolumn{2}{c}{\textbf{CVC-ClinicDB}}         & \multicolumn{2}{c}{\textbf{Kvasir-Seg}}           & \multicolumn{2}{c|}{\textbf{Synapse}}              \\ \hline
\multicolumn{1}{|c|}{Model}                       & \multicolumn{1}{l|}{Params} & \multicolumn{1}{l}{DSC (\%)} & \multicolumn{1}{l}{IoU (\%)} & \multicolumn{1}{l}{DSC (\%)} & \multicolumn{1}{l}{IoU (\%)} & \multicolumn{1}{l}{DSC (\%)} & \multicolumn{1}{l}{IoU (\%)} & \multicolumn{1}{l}{DSC (\%)} & \multicolumn{1}{l}{IoU (\%)} & \multicolumn{1}{l}{DSC (\%)} & \multicolumn{1}{l|}{IoU (\%)} \\ \hline
\multicolumn{1}{|c|}{U-Net}                       & \multicolumn{1}{c|}{8.64M}  & 88.87                  & 79.98                  & 77.14                  & 62.79                  & \textbf{89.63}         & \textbf{81.43}         & 82.99                  & 71.01                  & 78.55                  & 67.37                   \\
\multicolumn{1}{|c|}{\textbf{U-Net (DCA)}}        & \multicolumn{1}{c|}{8.75M}  & \textbf{89.66}         & \textbf{81.29}         & \textbf{78.13}         & \textbf{64.11}         & 89.53                  & 81.28                  & \textbf{84.03}         & \textbf{72.53}         & \textbf{78.98}         & \textbf{67.97}          \\
\multicolumn{1}{|c|}{ResUnet++}                   & \multicolumn{1}{c|}{13.1M}  & 85.43                  & 74.62                  & 75.68                  & 60.87                  & 89.46                  & 81.14                  & \textbf{82.26}         & \textbf{69.93}         & 75.91                  & 64.61                   \\
\multicolumn{1}{|c|}{\textbf{ResUnet++ (DCA)}}    & \multicolumn{1}{c|}{13.1M}  & \textbf{87.35}         & \textbf{77.56}         & \textbf{77.40}         & \textbf{63.13}         & \textbf{90.19}         & \textbf{82.32}         & 82.07                  & 69.74                  & \textbf{77.35}         & \textbf{66.43}          \\
\multicolumn{1}{|c|}{MultiResUnet}                & \multicolumn{1}{c|}{7.24M}  & \textbf{88.99}         & \textbf{80.18}         & 76.99                  & 62.59                  & 89.52                  & 81.35                  & 81.34                  & 68.66                  & 78.12                  & 67.30                   \\
\multicolumn{1}{|c|}{\textbf{MultiResUnet (DCA)}} & \multicolumn{1}{c|}{7.35M}  & 88.86                  & 79.98                  & \textbf{78.52}         & \textbf{64.63}         & \textbf{89.95}         & \textbf{81.91}         & \textbf{82.32}         & \textbf{70.00}         & \textbf{79.50}         & \textbf{68.65}          \\
\multicolumn{1}{|c|}{R2Unet}                      & \multicolumn{1}{c|}{9.78M}  & 85.16                  & 74.26                  & 78.20                  & 64.20                  & 88.12                  & 78.88                  & 81.07                  & 68.28                  & 75.86                  & 63.94                   \\
\multicolumn{1}{|c|}{\textbf{R2Unet (DCA)}}       & \multicolumn{1}{c|}{9.89M}  & \textbf{87.21}         & \textbf{77.37}         & \textbf{78.52}         & \textbf{64.64}         & \textbf{88.39}         & \textbf{79.28}         & \textbf{82.19}         & \textbf{69.89}         & \textbf{75.90}         & \textbf{64.85}          \\
\multicolumn{1}{|c|}{V-Net}                       & \multicolumn{1}{c|}{35.97M} & 88.78                  & 79.85                  & 74.79                  & 59.74                  & 88.09                  & 79.02                  & 80.79                  & 68.07                  & 79.27                  & 68.58                   \\
\multicolumn{1}{|c|}{\textbf{V-Net (DCA)}}        & \multicolumn{1}{c|}{36.08M} & \textbf{89.03}         & \textbf{80.27}         & \textbf{77.53}         & \textbf{63.31}         & \textbf{89.46}         & \textbf{81.07}         & \textbf{81.92}         & \textbf{69.53}         & \textbf{79.58}         & \textbf{69.00}          \\
\multicolumn{1}{|c|}{DoubleUnet}                  & \multicolumn{1}{c|}{29.68M} & 89.07                  & 80.30                   & 77.16                  & 62.82                  & 90.20                  & 82.35                  & 84.40                  & 73.08                  & 79.76                  & 69.31                   \\
\multicolumn{1}{|c|}{\textbf{DoubleUnet (DCA)}}   & \multicolumn{1}{c|}{30.68M} & \textbf{89.90}         & \textbf{81.68}         & \textbf{79.50}         & \textbf{65.97}         & \textbf{90.86}         & \textbf{83.47}         & \textbf{85.16}         & \textbf{74.34}         & \textbf{80.22}         & \textbf{69.80}          \\ \hline
\end{tabular}
\label{table1}
\end{center}
\caption{\label{table1} Performance comparison for plain and DCA integrated models on different datasets using DSC and IoU metrics. Boldfaced results indicate better results.}
\end{table*}

\subsection{Spatial Cross-Attention (SCA)}
SCA module is illustrated in Fig.\ref{fig:2}c. Given the reshaped outputs  $\pmb{\Bar{\text{T}}}_\text{i} \in \mathbb{R} ^ {\text{P}\times\text{C}_\text{i}}$, (i = 1, 2, ..., n) of the CCA module, we perform layer normalization and concatenation along the channel dimension. Unlike the CCA module, we utilize concatenated tokens $\pmb{\Bar{\text{T}}}_\text{c}$ as queries and keys while we use each token $\pmb{\Bar{\text{T}}}_\text{i}$ as values. We utilize $1\times1$ depth-wise projection over the queries, keys, and values:


\begin{equation}
        \textbf{Q} = \text{DConv1D}_{\textbf{Q}}(\Bar{\textbf{T}}_\text{c}), 
\end{equation}
\begin{equation}
    \textbf{K} = \text{DConv1D}_{\textbf{K}}(\Bar{\textbf{T}}_\text{c}), 
\end{equation}
\begin{equation}
    \textbf{V}_\text{i} = \text{DConv1D}_{\textbf{V}_\text{i}}(\Bar{\textbf{T}}_\text{i})
\end{equation}

where $\textbf{Q} \in \mathbb{R} ^ {\text{P}\times\text{C}_\text{c}}$, $\textbf{K} \in \mathbb{R} ^ {\text{P}\times\text{C}_\text{c}}$, $\textbf{V}_\text{i} \in \mathbb{R} ^ {\text{P}\times\text{C}_\text{i}}$ are the projected queries, keys and values, respectively. Then, SCA can be expressed as: 

\begin{equation}
\text{SCA}(\textbf{Q}, \textbf{K}, \textbf{V}_\text{i}) = \text{Softmax}\left(\frac{{\textbf{Q}}\textbf{K}^T}{\sqrt{d_k}}\right)\textbf{V}_\text{i}
\end{equation}

Here, $\textbf{Q}$, $\textbf{K}$, and $\textbf{V}_\text{i}$ are matrices representing the query, key, and value embeddings, respectively, and $\frac{1}{\sqrt{d_k}}$ is the scaling factor. For the multi-head case $d_k = \frac{\textbf{C}_\text{c}}{\text{h}_\text{c}}$ where $\text{h}_\text{c}$ is the number of heads. Outputs of the SCA module are then projected using depth-wise convolutions to form DCA outputs. Then, we apply layer normalization and GeLU to those DCA outputs. Finally, n outputs of the DCA block are connected to their corresponding decoder parts by up-sample layers followed by $1\times1$ convolution, batch normalization \cite{ioffe2015batch} and a ReLU \cite{nair2010rectified} sequence. Note that the major difference between cross-attention and self-attention is that cross-attention creates attention maps by fusing multi-scale encoder features together rather than utilizing each stage individually which also allows the cross-attention to capture long-range dependencies between different stages of the encoder.

\section{Experiments}
\label{sec:experiments}
\subsection{Datasets}
We conduct our experiments with five benchmark medical image segmentation datasets, including GlaS \cite{sirinukunwattana2017gland}, MoNuSeg \cite{kumar2017dataset}, CVC-ClinicDB \cite{bernal2015wm}, Kvasir-SEG \cite{jha2020kvasir} and Synapse. GlaS is a Gland Segmentation dataset consisting of 85 images for training and 80 images for testing. MoNuSeg is a nuclear segmentation dataset for digital microscopic tissue images. It includes 30 images for training and 14 images for testing. CVC-ClinicDB is a colonoscopy image dataset that includes a total of 612 images with their annotations. Kvasir-SEG is a polyp segmentation dataset that has 1000 annotated images. Following \cite{jha2019resunet++}, we randomly split 80\% of CVC-ClinicDB and Kvasir-Seg datasets into training and 20\% for testing. Synapse is a multi-organ segmentation dataset, consisting of 30 abdominal CT scans in 8 abdominal organs (aorta, gallbladder, spleen, left kidney, right kidney, liver, pancreas, spleen, stomach). Following \cite{chen2021transunet} and \cite{wang2022uctransnet}, we randomly choose 18 of these scans for training and 12 for testing.

\subsection{Models}
We test our DCA mechanism using six models with skip-connections including U-Net \cite{ronneberger2015u}, V-Net \cite{milletari2016v}, R2Unet \cite{alom2018recurrent}, ResUnet++ \cite{jha2019resunet++}, DoubleUnet \cite{jha2020doubleu} and MultiResUnet \cite{ibtehaz2020multiresunet}. For U-Net, V-Net, R2Unet and ResUnet++, we simply place our DCA block between the encoder and decoder stages as shown in Fig.\ref{fig:1}. For MultiResUnet, we place our DCA block at the end of the residual paths to improve the residual path outputs. DoubleUnet has three skip-connection schemes since two U-Net architectures are stacked on top of each other. The first scheme connects the first encoder (VGG19) to the first decoder, the second scheme connects the first encoder features to the second decoder and the last scheme connects the second encoder features to the second decoder. In our experiments, we integrated three DCA blocks into all those three skip-connection schemes.
\begin{figure*}[t]
\centering
\includegraphics[width=17.6cm, height=16.0cm]{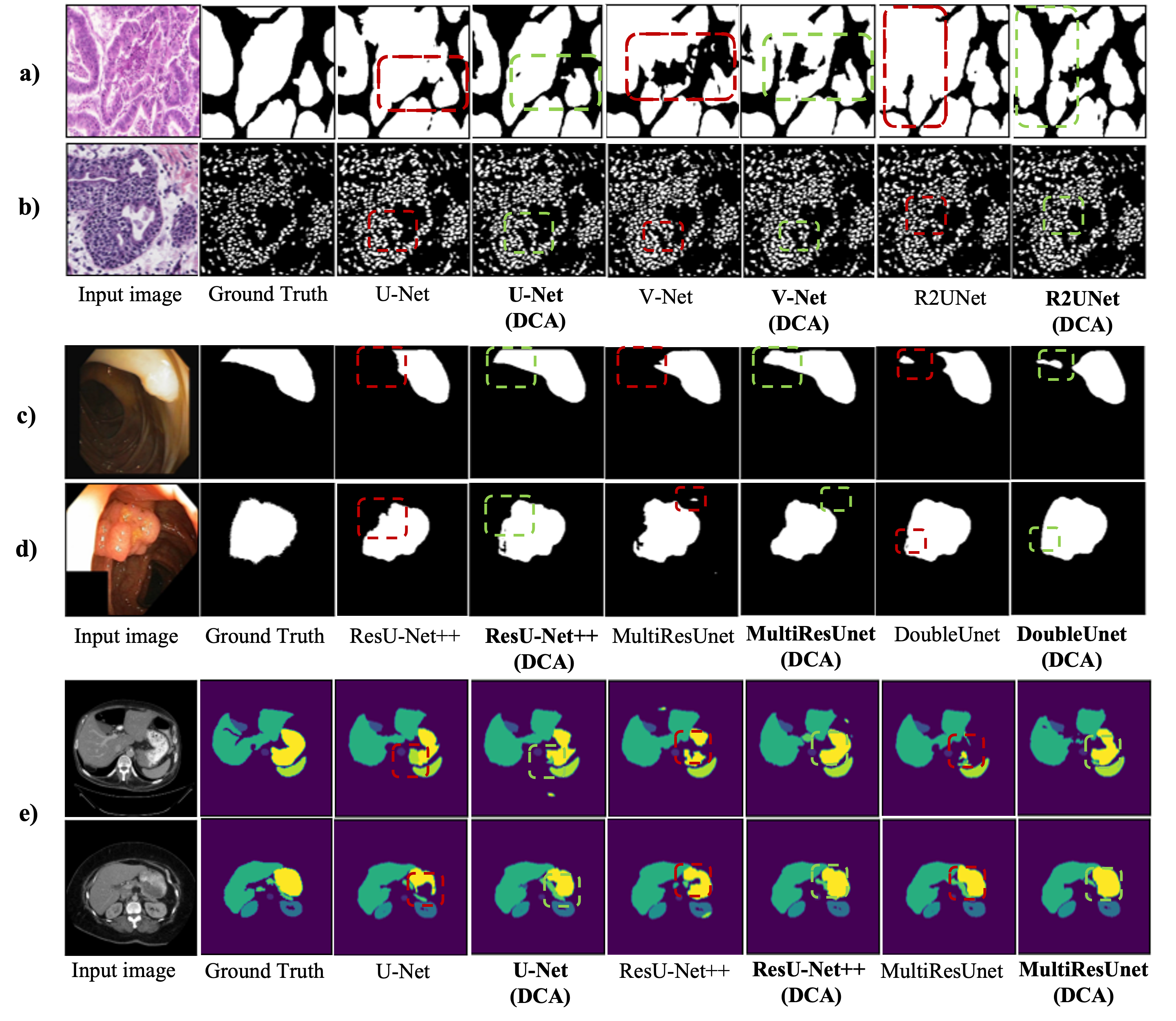}\\
\caption{Visual comparison for plain and DCA integrated models. (\textbf{a)} Glas, \textbf{b)} MoNuSeg, \textbf{c)} CVC-ClinicDB, \textbf{d)} Kvasir-Seg, \textbf{e)} Syanpse)}
\label{fig:3} 
\end{figure*}

\subsection{Implementation Details}
We implement our method in Pytorch using two NVIDIA Tesla V100-SXM2-16GB GPUs. We resize images to $224\times224$ for all datasets \cite{chen2021transunet, wang2022uctransnet, cao2023swin}. We set the patch size to 4 for ResUnet++ and 8 for the remaining models. We set the number of heads in CCA to 1 while we set it to 4 for SCA. As for the data augmentation, we perform random rotations and random vertical and horizontal flips. Following \cite{valanarasu2021medical} and \cite{wang2022uctransnet} we set the batch size to 4 for GlaS and MoNuSeg datasets while we set the batch size to 24 for Synapse dataset \cite{chen2021transunet, wang2022uctransnet}. For CVC-ClinicDB and Kvasir-Seg datasets, we set the batch size to 16 following \cite{jha2020doubleu, dong2021polyp}. We used Adam optimizer \cite{kingma2014adam} with an initial learning rate of $10^{-4}$ for all models. We utilized Dice loss \cite{sudre2017generalised} as the loss function while we used Dice Similarity Coefficient (DSC) and Intersection over Union (IoU) as the performance metrics. We trained all models for 200 epochs and reported the best results for each model.

\subsection{Results}
We verify the effectiveness of our proposed DCA block by conducting extensive experiments using six U-Net-based models and five benchmark medical image segmentation datasets. For a fair comparison, we use the same training settings for plain and DCA-integrated models. The overall results are shared in Table \ref{table1}. As mentioned before, we compare each model with their DCA integrated forms using DSC and IoU metrics where bold-faced results denote better performance. As can be seen in Table \ref{table1}, DCA can significantly improve the segmentation performance at a slight parameter increase for each model.  Note that the parameter increase in each model with DCA depends on the total number of skip-connection layers. ResUnet++ consists of three skip-connection layers which bring less than 0.7\% parameter increase with DCA while for U-Net, MultiResUnet, R2Unet and V-Net all of which include four skip-connection layers, the parameter increase varies between 0.3\% and 1.5\%, depending on the model capacity. Since DoubleUnet consists of three skip-connection schemes, each having four skip-connections, the total parameter increase is 3.4\% which is still a small increase. Considering such small additional parameters, DCA can provide DSC improvements up to 2.05\% on GlaS, 2.74\% on MoNuSeg, 1.37\% on CVC-ClinicDB, 1.12\%  on Kvasir-Seg and 1.44\% on Synapse datasets which show the effectiveness of our DCA mechanism.  

We also visually compare the model predictions to further verify our DCA mechanism. Fig.\ref{fig:3} depicts some of the model segmentation predictions. Red dashed rectangles show the regions where plain models struggled to provide accurate predictions while green dashed rectangles show the improvement of the DCA block in those same regions. As can be seen in Fig.\ref{fig:3}, models with DCA mechanism outperform the plain models by providing more consistent boundaries and preserving accurate shape information. Besides, models with DCA can better distinguish discrete parts by eliminating false positive predictions.
\begin{table}[t]
\scriptsize
\renewcommand\arraystretch{1.3}
\begin{center}
\begin{tabular}{|cccccc|}
\hline
\multicolumn{2}{|l}{}                                               & \multicolumn{2}{c}{\textbf{GlaS}} & \multicolumn{2}{c|}{\textbf{MoNuSeg}} \\ \hline
\multicolumn{1}{|c|}{Model}           & \multicolumn{1}{c|}{Params} & DSC (\%)             & IoU (\%)             & DSC (\%)               & IoU (\%)               \\ \hline
\multicolumn{1}{|c|}{U-Net}           & \multicolumn{1}{c|}{8.64M}  & 88.87          & 79.98          & 77.14            & 62.79            \\
\multicolumn{1}{|c|}{U-Net (CCA)}     & \multicolumn{1}{c|}{8.74M}  & 89.07          & 80.31          & 77.78            & 63.64            \\
\multicolumn{1}{|c|}{U-Net (SCA)}     & \multicolumn{1}{c|}{8.74M}  & 89.48          & 80.96          & 77.36            & 63.07            \\
\multicolumn{1}{|c|}{U-Net (SCA-CCA)} & \multicolumn{1}{c|}{8.75M}  & 89.03          & 80.24          & 77.90             & 63.80             \\
\multicolumn{1}{|c|}{U-Net (CCA-SCA)} & \multicolumn{1}{c|}{8.75M}  & \textbf{89.66} & \textbf{81.29} & \textbf{78.13}   & \textbf{64.11}   \\ \hline
\end{tabular}
\label{table2}
\end{center}
\caption{\label{table2} Quantitative comparison of different DCA layouts.}
\end{table}

\begin{table}[t]
\scriptsize
\renewcommand\arraystretch{1.3}
\begin{center}
\begin{tabular}{|cccccc|}
\hline
\multicolumn{2}{|c}{}                                                & \multicolumn{2}{c}{\textbf{GlaS}}          & \multicolumn{2}{c|}{\textbf{MoNuSeg}}      \\ \hline
\multicolumn{1}{|c|}{Model}            & \multicolumn{1}{c|}{Params} & DSC (\%)             & IoU (\%)             & DSC (\%)             & IoU (\%)             \\ \hline
\multicolumn{1}{|c|}{U-Net}            & \multicolumn{1}{c|}{8.75M}  & 88.87          & 79.98          & 77.14          & 62.79          \\
\multicolumn{1}{|c|}{U-Net (CCA+SCA)}  & \multicolumn{1}{c|}{8.75M}  & 89.09          & 80.38          & 77.30          & 63.00          \\
\multicolumn{1}{|c|}{U-Net (CCA\(\vert \vert\)SCA)} & \multicolumn{1}{c|}{8.75M}  & 89.19          & 80.52          & 78.03          & 63.97          \\
\multicolumn{1}{|c|}{U-Net (CCA-SCA)}  & \multicolumn{1}{c|}{8.75M}  & \textbf{89.66} & \textbf{81.29} & \textbf{78.13} & \textbf{64.11} \\ \hline
\end{tabular}
\label{table3}
\end{center}
\caption{\label{table3} Quantitative comparison of different fusion strategies for CCA and SCA.}
\end{table}
\section{Ablation Study}
\label{sec:ablationstudy}

\subsection{DCA Layout}
We start our ablation study by searching for the best layout for the proposed DCA mechanism. We first perform CCA and SCA modules individually. As shown in Table. \ref{table2}, both CCA and SCA modules outperform U-Net on GlaS dataset by 0.2\% and 0.61\% DSC improvements, respectively. When we incorporate CCA and SCA modules by CCA-SCA sequence (i.e., CCA first), the performance further improves by 0.79\% whereas the SCA-CCA sequence only improves the performance by 0.16\%. As for the MoNuSeg dataset, CCA and SCA modules improve U-Net by 0.64\% and 0.22\%, respectively while the SCA-CCA sequence performs slightly better than both individual CCA and SCA modules. Nonetheless, the dual attention scheme with the CCA-SCA sequence provides the best performance on both datasets, showing that channel and spatial-wise cross-attention mechanisms complement each other.

\subsection{Fusion of CCA and SCA}
As mentioned before, several fusion strategies for dual attention schemes have been proposed and shown to be effective. We conduct experiments using three fusion strategies: (i) fusion by summation (performing CCA and SCA in a parallel manner and taking the sum of their outputs) \cite{fu2019dual, mou2021cs2}; (ii) fusion by concatenation (performing CCA and SCA in a parallel manner and concatenating their outputs) \cite{liu2021scsa}; and (iii) sequential fusion (performing CCA and SCA in a sequential manner) \cite{ding2022davit}. Table. \ref{table3} shows the comparison of those three fusion strategies where $+$, \(\vert \vert\) and $-$ denote for fusion by summation (i), fusion by concatenation (ii) and sequential fusion (iii), respectively. Although both summation and concatenation fusion strategies improve U-Net on both datasets, the sequential fusion scheme performs the best results.

\begin{table}[t]
\scriptsize
\renewcommand\arraystretch{1.4}
\begin{center}
\begin{tabular}{|cccccc|}
\hline
\multicolumn{2}{|c}{}                                                & \multicolumn{2}{c}{\textbf{GlaS}}          & \multicolumn{2}{c|}{\textbf{MoNuSeg}}      \\ \hline
\multicolumn{1}{|c|}{Model}            & \multicolumn{1}{c|}{Params} & DSC (\%)             & IoU (\%)             & DSC (\%)             & IoU (\%)             \\ \hline
\multicolumn{1}{|c|}{U-Net}            & \multicolumn{1}{c|}{8.64M}  & 88.87          & 79.98          & 77.14          & 62.79          \\
\multicolumn{1}{|c|}{U-Net (DCA-Conv)} & \multicolumn{1}{c|}{9.01M}  & 89.52          & 81.07          & 77.62          & 63.43          \\
\multicolumn{1}{|c|}{U-Net (DCA-AP)}   & \multicolumn{1}{c|}{8.75M}  & \textbf{89.66} & \textbf{81.29} & \textbf{78.13} & \textbf{64.11} \\ \hline
\end{tabular}
\label{table4}
\end{center}
\caption{\label{table4} Quantitative comparison of average pooling and convolution for patch embedding.}
\end{table}

\subsection{Average Pooling for Patch Embedding}
As for the last part of our ablation study, we compare simple 2D average pooling with convolutional patch embedding. As shown in Table. \ref{table4}, DCA with convolutional patch embedding strategy, although still significantly improving U-Net, performs slightly worse than DCA with 2D average pooling. Besides, for multi-scale encoder features $\pmb{\text{E}}_\text{i} \in \mathbb{R} ^ {\text{C}_\text{i}\times\frac{\text{H}}{\text{$2^\text{i-1}$}}\times\frac{\text{W}}{\text{$2^\text{i-1}$}}}$, convolutional patch embedding requires kernel sizes ($\text{P}^s_\text{i}$, $\text{P}^s_\text{i}$) where $\text{P}^s_\text{i} = \frac{P^s}{2^{i-1}}$, (i = 1, 2, ..., n) which brings additional parameters ($\approx 260\text{K}$) while 2D average pooling operation is parameter-free and performs better when combined with $1\times1$ depth-wise convolutional projections.

\section{Conclusion}
\label{sec:conclusion}
In this paper, we introduced Dual Cross-Attention (DCA) to strengthen skip-connections in U-Net-based architectures for medical image segmentation. DCA consists of Channel Cross-Attention (CCA) and Spatial Cross-Attention (SCA) modules which sequentially capture long-range dependencies in channel and spatial dimensions, respectively. Besides, DCA utilizes cross-attention in order to effectively fuse low-level multi-scale encoder features and extract fine-grained representations to narrow the semantic gap. Our DCA mechanism is formed with lightweight operations such as 2D average pooling for patch embedding and depth-wise convolutions for projection layers to minimize the computational overhead. Comprehensive experiments using six state-of-the-art U-Net-based architectures and five benchmark medical image segmentation datasets show that DCA can significantly improve the segmentation performance for models with skip-connections.

\section*{Acknowledgment}
 We would like to acknowledge the Engineering Cancer Cures funding and IDSC Early Career Researcher Grant at the University of Miami for supporting this project.  
 We also thank the University of Miami Institute for Data Science and Computing for allocating two NVIDIA Tesla V100-SXM2-16GB GPUs for our experiments.

{\small
\bibliographystyle{ieee_fullname}
\bibliography{egbib}
}

\end{document}